\pdfoutput=1

\documentclass[11pt]{article}

\usepackage[]{EMNLP2023}

\usepackage{times}
\usepackage{latexsym}

\usepackage[T1]{fontenc}

\usepackage[utf8]{inputenc}

\usepackage{microtype}

\usepackage{inconsolata}
\usepackage{booktabs}
\usepackage{enumitem}
\usepackage{fdsymbol}
\usepackage{graphicx}
\usepackage{mathtools}
\usepackage{microtype}
\usepackage{multirow}
\usepackage{subfigure}
\usepackage{xspace}

\usepackage{svg}
\usepackage{pgfplots}
\usepackage{adjustbox}
\usepgfplotslibrary{external}
\usepgfplotslibrary{colorbrewer}
\pgfplotsset{
    cycle list/Dark2,
    cycle multiindex* list={
        mark list*\nextlist
        Dark2\nextlist
    },
    every axis plot/.append style={line width=1pt},
}
\pgfplotsset{compat=newest}
\usetikzlibrary{shapes.geometric}

\newcommand\blfootnote[1]{%
  \begingroup
  \renewcommand\thefootnote{}\footnote{#1}%
  \addtocounter{footnote}{-1}%
  \endgroup
}

\title{Smaller Language Models are capable of selecting Instruction-Tuning \\ Training Data for Larger Language Models}

\author{
  Dheeraj Mekala$^{*, \diamondsuit}$ $\quad$ Alex Nguyen$^{*, \diamondsuit}$ $\quad$ Jingbo Shang$^{\heartsuit, \diamondsuit, \spadesuit}$ $\quad$ \\
  $^\diamondsuit$Department of Computer Science \& Engineering, University of California San Diego\\
  $^\heartsuit$ Hal\i c\i o\u glu Data Science Institute, University of California San Diego\\
  \small \texttt{\{dmekala, atn021, jshang\}@ucsd.edu}
}

\begin{document}
\maketitle

\begin{abstract}
    Instruction-tuning language models has become a crucial step in aligning them for general use. 
    Typically, this process involves extensive training on large datasets, incurring high training costs.
    In this paper, we introduce a novel training data selection based on the learning percentage of the samples.
    We assert that current language models possess the capability to autonomously select high-quality training data, leading to comparable or improved performance compared to training on the entire dataset. 
    Our experiments span different-sized models, revealing that this characteristic holds for models ranging from 1B (small) to 13B (large) in size. 
    Moreover, we demonstrate an interesting finding that the data hardness transfers across model sizes, and a smaller 350M model can effectively curate high-quality training data with hard samples for a larger 13B model, resulting in an equally or superior instruction-tuned model compared to training on the complete dataset. 
    Utilizing open-sourced OPT and Llama-2 models up to 13B in size, two publicly available instruction-tuning training datasets and evaluated by both automatic metrics \& humans, our paper introduces a novel approach to training data selection, showcasing a more efficient alternative.

    \blfootnote{$*$ Equal Contribution}
    \blfootnote{$\spadesuit$ Corresponding Author}
\end{abstract}
\section{Introduction}
Instruction tuning empowers large language models (LLMs) to generalize to novel tasks and instills an instruction-following characteristic, marking the initial stride towards aligning them for general use~\cite{Sanh2021MultitaskPT, Ye2022GuessTI, Wei2021FinetunedLM, Chung2022ScalingIL, Ye2022GuessTI}. 
This process involves fine-tuning language models with extensive sets of real~\cite{Mishra2021CrossTaskGV, Wang2022SuperNaturalInstructionsGV} and/or synthetic instructions~\cite{Wang2022SelfInstructAL, Honovich2022UnnaturalIT}. 
Given that these datasets are typically vast, encompassing thousands of samples, the training costs associated with this approach are notably high.

\begin{figure}[t]
\centering
\begin{adjustbox}{max width=0.42\textwidth}
\begin{tikzpicture}
\begin{axis}[
    xmin=1, xmax=4, xtick={1,2,3,4}, x tick style={draw=none},
    xlabel=Ranking Source,
    xticklabels={350M, 1.3B, 2.7B, 6.7B},
    ybar, ymin=50, ymax=54,
    nodes near coords,
    bar width=32pt,
    bar shift=0pt,
    ylabel=Win rate,
    enlarge y limits=0.1,
    enlarge x limits=0.2,
]
\addplot coordinates {
    (1,51.74)
};
\addplot coordinates {
    (2,53.21)
};
\addplot coordinates {
    (3,53.68)
};
\addplot coordinates {
    (4,53.34)
};
\end{axis}
\end{tikzpicture}
\end{adjustbox}
\caption{The win rate of OPT-13B model trained on 10\% data sub-sampled by smaller OPT models (350M, 1.3B, 2.7B)
from Alpaca Data, is compared against the OPT-13B model trained on the full dataset. All win rates exceed 50, indicating even a smaller 350M dataset can curate high-quality data for a larger 13B model.}
\label{fig:overview}
\end{figure}
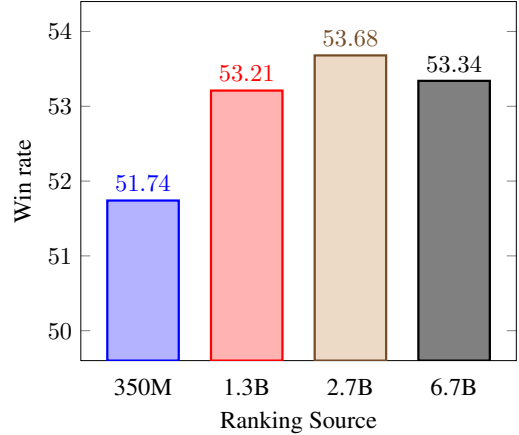

Past research into the memorization effects of deep neural networks have revealed a tendency to memorize easy instances first and gradually learn more challenging instances towards the end~\cite{arpit2017closer, geifman2018bias, zhang2021understanding, mekala-etal-2022-lops}. 
Additionally, \cite{swayamdipta-etal-2020-dataset} show that ambiguous and hard samples in training data are sufficient for achieving good generalization.
The process of data selection, wherein subsets are chosen from extensive training data to achieve superior performance, has attracted significant attention among researchers recently. 
Earlier studies involved manual feature engineering of various indicators from the data~\cite{Cao2023InstructionMH}, training large custom models~\cite{Li2023FromQT}, or employing closed LLMs like GPT-3.5~\cite{chen2023alpagasus} for data selection.

In this paper, we delve into the measurement of sample difficulty from the model's perspective. 
Drawing inspiration from the learning order metric in \cite{mekala-etal-2022-lops}, we propose a novel data selection method that utilizes the learning percentage as a difficulty metric that the model can use to self-rank its training data. 
Essentially, the more learning that occurs in earlier epochs, the easier the sample is considered. 
We then select the most difficult sample subsets based on this ranking and instruction-tune a language model. 
Our experiments involve two instruction-tuning datasets, Alpaca-Data~\cite{alpaca}, and Dolly~\cite{DatabricksBlog2023DollyV2}, with performance measured using automated metrics such as AlpacaEval~\cite{alpaca_eval} and human evaluation.

Our main findings indicate that language models can autonomously select training data, achieving performance equal to or better than training on the entire dataset. 
Furthermore, this characteristic scales across different model sizes, ranging from smaller ones (1B) to larger ones (13B)\footnote{Due to limitations in our compute, the largest size we were able to train is 13B.} in parameters. 
As the size of the language model increases, we observe a consistent reduction in the minimum amount of data needed to surpass the performance of a model trained on the entire dataset.
Interestingly, we observe that the data hardness also transfers across models, meaning samples considered difficult by smaller models are similarly challenging for larger models.
Moreover, we note that this transferability improves with the size of the smaller model, eventually achieving comparable quality, beyond a size threshold, to that attained by self-selection conducted by larger models.
Our study employs open-sourced models such as OPT~\cite{Zhang2022OPTOP} and Llama-2~\cite{Touvron2023Llama2O} to support these findings.

The remainder of the paper is structured as follows: initially, we describe the experimental setup encompassing the language models, the datasets employed, and the evaluation metrics utilized (\autoref{sec:exp_setup}). 
Subsequently, we present our learning percentage-based difficulty metric and analyze it in detail (\autoref{sec:lp}). 
Following this, we optimize the proposed metric and introduce an equally effective, approximate, and faster metric (\autoref{sec:lp_app}). 
Ultimately, we analyze the challenging data identified through this metric (\autoref{sub:dissect}).

We publicly release the code here\footnote{\url{https://github.com/dheeraj7596/Small2Large}}.

\section{Experiment Setup}
\label{sec:exp_setup}
We design controlled experiments to empirically validate our assertions. 
Our experimental setup encompasses language models spanning various families and sizes, alongside multiple datasets, the specifics of which are detailed below. 

\subsection{Language Models \& Evaluation}
We use OPT (1.3B, 2.7B, 6.7B, 13B) and Llama-2 (7B, 13B) for experiments.
We fine-tune all models for three epochs on three NVIDIA A100 GPUs.
For the comparison of language models, we employ AlpacaEval—an automated evaluator that tasks a larger language model with selecting the superior response from two LMs. 
AlpacaEval offers an evaluation set comprising 805 samples, designed to assess general instruction-following capabilities by combining data from various sources, including self-instruct~\cite{Wang2022SelfInstructAL}, anthropic helpfulness\footnote{\url{https://huggingface.co/datasets/Anthropic/hh-rlhf/viewer/Anthropic--hh-rlhf/test}}, open assistant~\cite{Kopf2023OpenAssistantC}, Koala\footnote{\url{https://github.com/arnav-gudibande/koala-test-set}}, and Vicuna~\cite{vicuna2023} evaluation sets.
While AlpacaEval provides various options for the judge language model, we opt for GPT-3.5 (gpt-3.5-turbo-16k-0613)~\cite{chatgpt} due to its cost-effectiveness.

\subsection{Data}
We experiment on Alpaca-Data~\cite{alpaca} and Dolly~\cite{DatabricksBlog2023DollyV2} datasets.
Alpaca-Data comprises 52,000 samples generated through the self-instruct method by prompting text-davinci-003 with 175 human-written seed instruction-output pairs~\cite{Wang2022SelfInstructAL}.
Dolly, on the other hand, consists of 15,000 human-generated samples. 



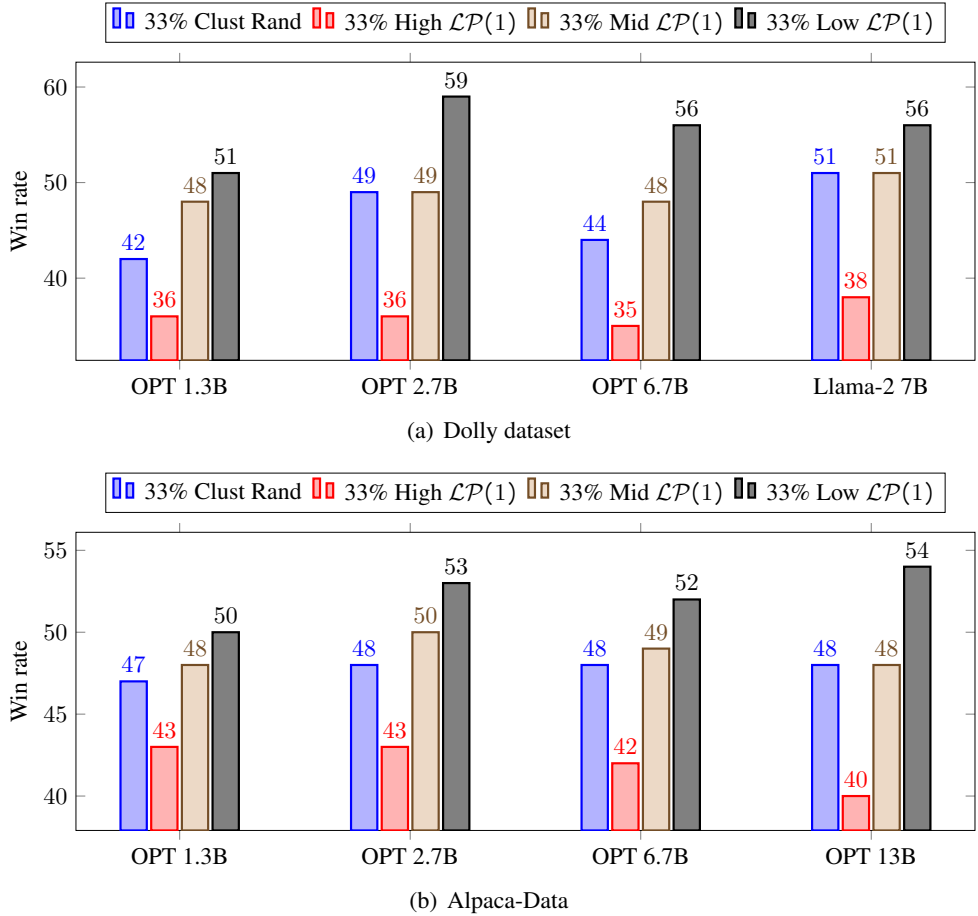
\begin{figure*}[t]
    \centering
    \subfigure[Dolly dataset]{
        \begin{adjustbox}{max width=0.8\textwidth}
            \begin{tikzpicture}
                \begin{axis}[
                    symbolic x coords={OPT 1.3B,OPT 2.7B,OPT 6.7B, Llama-2 7B},
                    xtick=data,
                    width=\textwidth,
                    height=0.4\textwidth,
                    nodes near coords,
                    ybar,
                    bar width=12pt,
                    legend style={at={(0.5,1.2)}, anchor=north, legend columns=-1, legend cell align={left}, column sep=0.25em},
                    ylabel=Win rate,
                    enlarge y limits=0.15,
                    enlarge x limits=0.15, 
                ]
                \addplot coordinates {
                    (OPT 1.3B,42)
                    (OPT 2.7B,49)
                    (OPT 6.7B,44)
                    (Llama-2 7B, 51)
                };
                \addplot coordinates {
                    (OPT 1.3B,36)
                    (OPT 2.7B,36)
                    (OPT 6.7B,35)
                    (Llama-2 7B,38)
                };
                \addplot coordinates {
                    (OPT 1.3B,48)
                    (OPT 2.7B,49)
                    (OPT 6.7B,48)
                    (Llama-2 7B,51)
                };
                \addplot coordinates {
                    (OPT 1.3B,51)
                    (OPT 2.7B,59)
                    (OPT 6.7B,56)
                    (Llama-2 7B,56)
                };
                \legend{33\% Clust Rand, 33\% High $\mathcal{LP}(1)$, 33\% Mid $\mathcal{LP}(1)$, 33\% Low $\mathcal{LP}(1)$}
                \end{axis}
            \end{tikzpicture}
        \end{adjustbox}
        \label{dolly_33}
    }
    \subfigure[Alpaca-Data]{
        \begin{adjustbox}{max width=0.8\textwidth}
            \begin{tikzpicture}
                \begin{axis}[
                    symbolic x coords={OPT 1.3B,OPT 2.7B,OPT 6.7B,OPT 13B},
                    xtick=data,
                    width=\textwidth,
                    height=0.4\textwidth,
                    nodes near coords,
                    ybar,
                    bar width=12pt,
                    legend style={at={(0.5,1.2)}, anchor=north, legend columns=-1, legend cell align={left}, column sep=0.25em},
                    ylabel=Win rate,
                    enlarge y limits=0.15,
                    enlarge x limits=0.15, 
                ]
                \addplot coordinates {
                    (OPT 1.3B,47)
                    (OPT 2.7B,48)
                    (OPT 6.7B,48)
                    (OPT 13B, 48)
                };
                \addplot coordinates {
                    (OPT 1.3B,43)
                    (OPT 2.7B,43)
                    (OPT 6.7B,42)
                    (OPT 13B,40)
                };
                \addplot coordinates {
                    (OPT 1.3B,48)
                    (OPT 2.7B,50)
                    (OPT 6.7B,49)
                    (OPT 13B,48)
                };
                \addplot coordinates {
                    (OPT 1.3B,50)
                    (OPT 2.7B,53)
                    (OPT 6.7B,52)
                    (OPT 13B,54)
                };
                \legend{33\% Clust Rand, 33\% High $\mathcal{LP}(1)$, 33\% Mid $\mathcal{LP}(1)$, 33\% Low $\mathcal{LP}(1)$}
                \end{axis}
            \end{tikzpicture}
        \end{adjustbox}
        \label{fig:opt_33_alpaca}
    }
    \caption{We partition datasets into three equal-sized buckets based on their $\mathcal{LP}(1)$ scores. We train a model per bucket and report its win rate against the one trained on the complete dataset. The model used to compute $\mathcal{LP}(1)$ scores and trained is depicted on the X-axis and the win rate on the Y-axis. We observe the model trained on the lowest $\mathcal{LP}(1)$ values (33\% Low $\mathcal{LP}(1)$) exhibits superior performance compared to the others.}
\end{figure*}

\section{$\mathcal{LP}$: Learning Percentage as a Difficulty Metric}
\label{sec:lp}
The concept of learning order~\cite{dong2021data, mekala-etal-2022-lops} is designed to assess the quality of a sample in the context of a weakly-supervised classification problem~\cite{mekala-etal-2022-leveraging, mekala-shang-2020-contextualized}. 
The learning order of a data point is defined as the epoch at which it is learned during training, precisely when the model's predicted label aligns with the given ground truth. 
To adapt this concept to the text generation problem, we introduce the notion of learning percentage.

For a data point after epoch-$i$, the learning percentage is defined as the percentage drop in perplexity during epoch-$i$ compared to the total drop in perplexity by the end of training. 
Assuming a language model is fine-tuned for $n$ epochs, with $\mathcal{P}_i$ denoting the perplexity of a sample at the end of epoch-$i$ and $\mathcal{P}_0$ indicating its perplexity at the beginning of training, the learning percentage $\mathcal{LP}(i)$ at the end of epoch-$i$ is mathematically defined as follows:
\begin{equation}\label{eqn:idil}
    \mathcal{LP}(i) = \frac{\mathcal{P}_{i-1} - \mathcal{P}_{i}}{\mathcal{P}_0 - \mathcal{P}_n}
\end{equation}

A higher learning percentage at earlier epochs indicates that majority of the learning occurs during the initial epochs. 
Given the deep neural models typically learn easier samples initially and progress to more challenging samples later~\cite{arpit2017closer, geifman2018bias, zhang2021understanding, mekala-etal-2022-lops}, a higher learning percentage in the early epochs implies easy-to-learn samples.
Since language models are known to learn most of the information in just one epoch~\cite{Komatsuzaki2019OneEI, Hoffmann2022TrainingCL, Zhang2022OPTOP, Touvron2023Llama2O}, we consider $\mathcal{LP}(1)$ to rank the training data.

The diversity of training data is a pivotal attribute for achieving high quality and optimal performance~\cite{sorscher2022beyond, Tirumala2023D4IL}. 
To enhance this diversity, we employ k-means clustering on the sentence embeddings generated by the all-MiniLM-L6-v2 model\footnote{\url{https://www.sbert.net}} on the entire training dataset, ensuring that each cluster contains a minimum average of 50 samples.
As a result, the Alpaca-Data yields 1000 clusters with an average of 52 samples per cluster, while the Dolly dataset yields 300 clusters with an average of 50 samples per cluster.
Subsequently, we rank the training data using $\mathcal{LP}(1)$ in ascending order and select the top-$k$\% of samples from each cluster, i.e., the samples that are learned the least in the first epoch.  




\begin{figure}[t]
\centering
\begin{adjustbox}{max width=0.42\textwidth}
\begin{tikzpicture}
\begin{axis}[
    symbolic x coords={Llama-2 7B,Llama-2 13B},
    xtick=data,
    nodes near coords,
    ybar,
    bar width=12pt,
    legend style={at={(0.5,1.28)}, anchor=north, legend columns=2, legend cell align={left}, column sep=0.25em},
    ybar,
    ylabel=Win rate,
    enlarge y limits=0.15,
    enlarge x limits=0.5, 
]
\addplot coordinates {
    (Llama-2 7B,50)
    (Llama-2 13B,50)
};
\addplot coordinates {
    (Llama-2 7B,41)
    (Llama-2 13B,41)
};
\addplot coordinates {
    (Llama-2 7B,53)
    (Llama-2 13B,52)
};
\addplot coordinates {
    (Llama-2 7B,57)
    (Llama-2 13B,55)
};
\legend{33\% Clust Rand, 33\% High $\mathcal{LP}(1)$, 33\% Mid $\mathcal{LP}(1)$, 33\% Low $\mathcal{LP}(1)$}
\end{axis}
\end{tikzpicture}
\end{adjustbox}
\caption{We partition Alpaca-Data into three equal-sized buckets based on their $\mathcal{LP}(1)$ scores. 
The model used to compute $\mathcal{LP}(1)$ scores and trained is on the X-axis and the win rate on the Y-axis. We observe the model trained on the lowest $\mathcal{LP}(1)$ values (33\% Low $\mathcal{LP}(1)$) exhibits superior performance.}
\label{fig:llama_33_alpaca}
\end{figure}
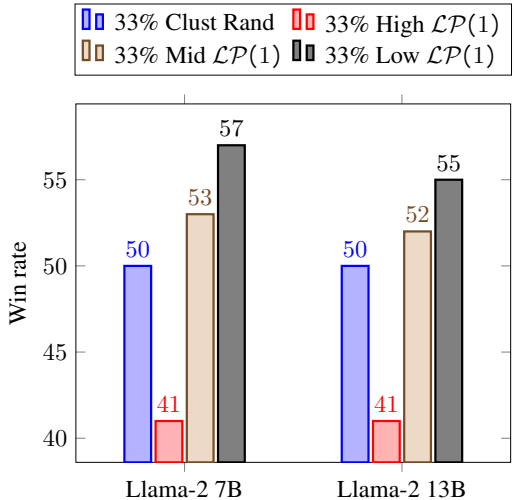



\subsection{$\mathcal{LP}(1)$-based Data Selection}
\label{sec:lp1}
We calculate the $\mathcal{LP}(1)$ scores of the training dataset and organize it in ascending order according to these scores.
Subsequently, we partition the dataset into three equal buckets. 
To enhance the diversity, this partitioning is done per cluster.
The bucket characterized by the lowest $\mathcal{LP}(1)$ values (\textbf{33\% Low $\mathcal{LP}(1)$}) represents the most challenging data in each cluster, while the bucket with the highest values corresponds to the least challenging data (\textbf{33\% High $\mathcal{LP}(1)$}) in each cluster.

We train one model per bucket and calculate the win rate against the model trained on the complete training dataset.
We also present the performance of the model trained on randomly selected 33\% data from each cluster (\textbf{33\% Clust Rand}) for reference.
The AlpacaEval win rate scores of the Llama-2 7B and 13B models trained on individual buckets in Alpaca-Data are plotted in Figure~\ref{fig:llama_33_alpaca}.
Similarly, the win rate scores of OPT 1.3B, 2.7B, 6.7B, and Llama-2 7B models on the Dolly dataset are shown in Figure~\ref{dolly_33}.
We observe that models trained on the bucket associated with the lowest $\mathcal{LP}(1)$ scores (33\% Low $\mathcal{LP}(1)$) consistently achieve scores exceeding 50, indicating that training on challenging samples alone is adequate for a robust instruction-tuning model. 
Furthermore, our analysis reveals that the model trained on the lowest $\mathcal{LP}(1)$ scores (33\% Low $\mathcal{LP}(1)$) consistently outperforms those trained on mid and high buckets by a significant margin. 
For example, in Figure~\ref{dolly_33}, OPT 2.7B trained on the low bucket outperforms the mid bucket by 10 points and the high bucket by 23 points.
This underscores a compelling argument that leveraging difficult data yields more favorable outcomes compared to training on easier datasets.
The 33\% High bucket results in worse performance than random selection for all 
models, highlighting that easy samples alone are insufficient.

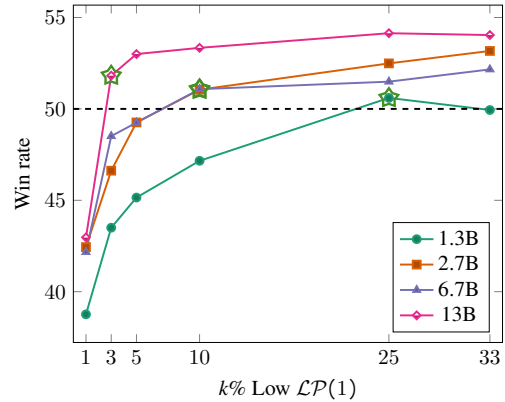
\begin{figure}[t]
\centering
\begin{adjustbox}{max width=0.42\textwidth}
\definecolor{dark-green}{HTML}{439023}
\pgfdeclareplotmark{mystar}{
    \node[star,star point ratio=1.65,minimum size=10pt,line width=1.2pt,inner sep=0pt,draw=dark-green,solid,fill=white,fill opacity=0] {};
}
\begin{tikzpicture}
\begin{axis}[
    xmin=0, xmax=34,
    width=0.58\textwidth,
    height=0.48\textwidth,
    xtick={1,3,5,10,25,33},
    ylabel=Win rate,
    xlabel=\textit{k}\% Low $\mathcal{LP}(1)$,
    legend pos=south east,
]
\addplot coordinates {
(1,38.76)(3,43.50)(5,45.15)(10,47.16)(25,50.60)(33,49.94)
};
\addplot coordinates {
(1,42.44)(3,46.62)(5,49.26)(10,51.05)(25,52.49)(33,53.17)
};
\addplot coordinates {
(1,42.17)(3,48.51)(5,49.25)(10,51.08)(25,51.49)(33,52.16)
};
\addplot coordinates {
(1,42.97)(3,51.82)(5,53.00)(10,53.34)(25,54.14)(33,54.04)
};
\addplot[only marks,mark=mystar] coordinates {
(3,51.82)(10,51.08)(10,51.05)(25,50.60)
};
\addplot[domain=0:34, dashed]{50};
\legend{1.3B,2.7B,6.7B,13B}
\end{axis}
\end{tikzpicture}
\end{adjustbox}
\caption{We consider Alpaca-Data, vary the percentage of data selected, and plot the win rate of OPT models, trained on the selected data in comparison to models trained on the complete dataset. The minimum percentage of data necessary for each model to surpass the 50\% threshold is highlighted with \protect\definecolor{dark-green}{HTML}{439023}\protect\tikz \protect\node[star,star point ratio=1.65,minimum size=8pt,line width=1.2pt,inner sep=0pt,draw=dark-green,solid,fill=white,fill opacity=0] {};.
}
\label{fig:scale_analysis}
\end{figure}

\paragraph{Performance vs Scale} 

To investigate the variation of this trait with scale, we plot the win rate scores of OPT (1.3B, 2.7B, 6.7B, 13B) models trained on each bucket within the Alpaca-Data in Figure~\ref{fig:opt_33_alpaca}. 
Remarkably, all models trained on the low bucket consistently achieve win rates exceeding 50, surpassing those of the corresponding mid and high buckets. 
This consistent trend across different model scales underscores the robustness of the observed pattern.

\paragraph{Larger models need fewer samples}
To further analyze the required amount of difficult data necessary for training high-quality instruction-tuned models of varying sizes, we analyze OPT (1.3B, 2.7B, 6.7B, 13B) models by varying the percentage of selected data in Alpaca-Data and plot their win rates against the corresponding models trained on the complete dataset in Figure~\ref{fig:scale_analysis}. 
Strikingly, we observe a downward trend in the minimum percentage of difficult data required (denoted by \protect\definecolor{dark-green}{HTML}{439023}\protect\tikz \protect\node[star,star point ratio=1.65,minimum size=8pt,line width=1.2pt,inner sep=0pt,draw=dark-green,solid,fill=white,fill opacity=0] {};) for achieving a win rate of at least 50 with an increase in the model's size. 
For example, the OPT-13B model outperforms its full dataset counterpart with only 3\% of the training data.
This suggests that as the model's size increases, the amount of challenging data required decreases, albeit the necessity for such difficult data persists.
This finding provides additional insight into the exceptional performance exhibited by the Llama-65B model when trained with 1000 difficult samples in \cite{Zhou2023LIMALI}.


\begin{figure*}[t]
    \centering
    \subfigure[Alpaca-Data, Llama models]{
        \begin{adjustbox}{max width=0.30\textwidth}
            \begin{tikzpicture}
                \begin{axis}[
                    symbolic x coords={Llama-2 7B, Self-Ranking (Llama-2 13B)},
                    xticklabels={
                        Llama-2 7B, Llama-2 13B\\(Self-Ranking)
                    },
                    xtick=data,
                    width=0.38\textwidth,
                    height=0.38\textwidth,
                    x tick label style={text width=4cm,align=center},
                    xlabel=$\mathcal{LP}(1)$ Ranking Source,
                    legend style={at={(0.5,1.35)}, anchor=north, legend columns=3, legend cell align={left}, column sep=0.25em, font=\Large},
                    ylabel=Win rate,
                    enlarge y limits=0.1,
                    enlarge x limits=0.1,
                    ticklabel style = {font=\Large},
                    label style = {font=\Large},
                ]
                \addplot coordinates {
                    (Llama-2 7B,56.62)
                    (Self-Ranking (Llama-2 13B),56.52)
                };
                \addplot coordinates {
                    (Llama-2 7B,55.92)
                    (Self-Ranking (Llama-2 13B),56.43)
                };
                \addplot coordinates {
                    (Llama-2 7B,57.23)
                    (Self-Ranking (Llama-2 13B),54.18)
                };
                \addplot coordinates {
                    (Llama-2 7B,57.74)
                    (Self-Ranking (Llama-2 13B),54.48)
                };
                \addplot coordinates {
                    (Llama-2 7B,56.15)
                    (Self-Ranking (Llama-2 13B),55.14)
                };
                \addplot coordinates {
                    (Llama-2 7B,53.22)
                    (Self-Ranking (Llama-2 13B),53.65)
                };
                \legend{1\% Low, 3\% Low, 5\% Low, 10\% Low, 25\% Low, 33\% Low}
                \end{axis}
            \end{tikzpicture}
        \end{adjustbox}
        \label{fig:small2large_llama_alpaca}
    }
    \subfigure[Dolly dataset, Llama models]{
        \begin{adjustbox}{max width=0.30\textwidth}
            \begin{tikzpicture}
                \begin{axis}[
                symbolic x coords={Llama-2 7B, Self-Ranking (Llama-2 13B)},
                xticklabels={
                    Llama-2 7B, Llama-2 13B\\(Self-Ranking)
                },
                xtick=data,
                width=0.38\textwidth,
                height=0.37\textwidth,
                x tick label style={text width=4cm,align=center},
                xlabel=$\mathcal{LP}(1)$ Ranking Source,
                legend style={at={(0.5,1.225)}, anchor=north, legend columns=3, legend cell align={left}, column sep=0.25em, font=\Large},
                ylabel=Win rate,
                enlarge y limits=0.1,
                enlarge x limits=0.1,
                ticklabel style = {font=\Large},
                label style = {font=\Large},
                ]
                \addplot coordinates {
                (Llama-2 7B,50.62)
                (Self-Ranking (Llama-2 13B),54.29)
                };
                \addplot coordinates {
                (Llama-2 7B,51.68)
                (Self-Ranking (Llama-2 13B),50.87)
                };
                \addplot coordinates {
                (Llama-2 7B,56.15)
                (Self-Ranking (Llama-2 13B),58.46)
                };
                \legend{3\% Low, 10\% Low, 33\% Low}
                \end{axis}
            \end{tikzpicture}
        \end{adjustbox}
        \label{fig:small2large_llama_dolly}
    }
    \subfigure[Alpaca-Data, OPT models]{
        \begin{adjustbox}{max width=0.35\textwidth}
            \begin{tikzpicture}
            \begin{axis}[
                symbolic x coords={350M,1.3B,2.7B,6.7B,13B (Self-Ranking)},
                xticklabels={
                    350M,
                    1.3B,
                    2.7B,
                    6.7B,
                    13B\\(Self-Ranking)
                },
                xtick=data,
                x tick label style={text width=4cm,align=center},
                xlabel=$\mathcal{LP}(1)$ Ranking Source,
                width=0.65\textwidth,
                height=0.43\textwidth,
                legend style={at={(0.5,1.33)}, anchor=north, legend columns=2, legend cell align={left}, column sep=0.25em, font=\LARGE},
                ylabel=Win rate,
                enlarge y limits=0.1,
                enlarge x limits=0.07,
                ticklabel style = {font=\LARGE},
                label style = {font=\LARGE}
            ]
            \addplot coordinates {
                (350M,50.97)
                (1.3B,50.58)
                (2.7B,52.44)
                (6.7B,52.03)
                (13B (Self-Ranking), 53.00)
            };
            \addplot coordinates {
                (350M,51.74)
                (1.3B,53.21)
                (2.7B,53.68)
                (6.7B,53.34)
                (13B (Self-Ranking), 54.04)
            };
            \addplot coordinates {
                (350M,53.44)
                (1.3B,54.31)
                (2.7B,55.28)
                (6.7B,54.14)
                (13B (Self-Ranking), 54.14)
            };
            \addplot coordinates {
                (350M,53.63)
                (1.3B,53.48)
                (2.7B,53.21)
                (6.7B,55.14)
                (13B (Self-Ranking), 54.04)
            };
            \legend{5\% Low, 10\% Low, 25\% Low, 33\% Low}
            \end{axis}
            \end{tikzpicture}
        \end{adjustbox}
        \label{fig:opt_small2large_alpaca}
    }
    \caption{
    We vary the percentage of selected data to train 13B model and conduct a comparison of win rates obtained when data is self-selected by the 13B model vs selected by smaller models. 
    The smaller model used is mentioned on the X-axis and the win rate is on the Y-axis.
    We observe that the data hardness transfers from smaller models to 13B, leading to improved or comparable performance compared to 13B model trained on the self-selected data.
    }
    \label{fig:all_small2large}
\end{figure*}
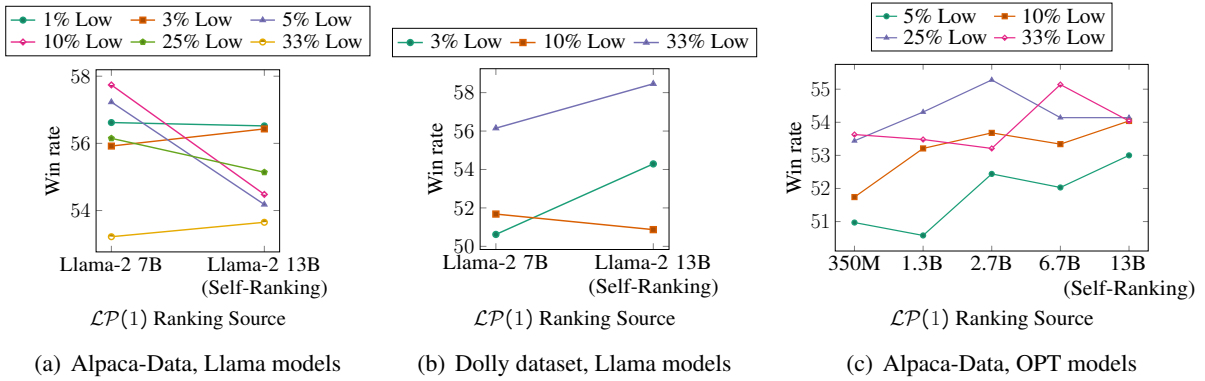

\subsection{Is Data Hardness transferable?}
\label{sec:transfer}
In the previous section, we observed challenging training data yields high-performing instruction-tuned models. 
In this section, we investigate transferability of data hardness, specifically whether samples deemed difficult by a smaller model are also considered difficult by a larger model.


To assess this, we consider a smaller and a larger model. 
We obtain $\mathcal{LP}(1)$ scores using the smaller model, following which we select the top-k\% (in ascending order) of training data based on these scores. 
Subsequently, we train the larger model using this selected dataset. 
For each experimental configuration, we calculate the win rate of the larger model trained on the selected data against it when trained on the complete dataset.

We consider Llama-2 7B as the smaller model and Llama-2 13B as the larger model. We vary the percentage data selected, and plot the win rate of Llama-2 13B trained on selected data in comparison to the one fine-tuned on the entire Alpaca-Data dataset in Figure~\ref{fig:small2large_llama_alpaca} and for Dolly dataset in Figure~\ref{fig:small2large_llama_dolly} respectively.
We find that the ranking of the Llama-2 7B model transfers effectively to the 13B model, resulting in a model of comparable or even improved quality in some instances. 
For example, in Figure~\ref{fig:small2large_llama_alpaca}, the win rate of the Llama-2-13B model trained on 10\% of the Alpaca-Data, selected by the 7B model, outperforms the self-ranking of the 13B model by 4 points.

Similarly, we consider OPT (1.3B, 2.7B, 6.7B) models as smaller models and OPT 13B as the larger model and plot the performance for the Alpaca-Data dataset in Figure~\ref{fig:opt_small2large_alpaca} and for the Dolly dataset in Figure~\ref{fig:small2large_opt_dolly} in Appendix~\ref{app:2} respectively.
From Figure~\ref{fig:opt_small2large_alpaca}, we observe that the performance of the 13B model increases with the size of the smaller model up to 2.7B and further plateaus where it eventually matches the self-selection performance of 13B model.
With only a 1.4-point drop in average win rate, even a small 350M model can be leveraged for curating training data for a large 13B model.
This demonstrates that the data hardness transfers efficiently from a smaller model to a larger one, improving with the size of the smaller model and eventually matching self-selection performance beyond a specific size threshold (2.7B).

\begin{figure}[t]
\centering
\begin{adjustbox}{width=0.42\textwidth}
\begin{tikzpicture}
\begin{axis}[
    symbolic x coords={350M, 1.3B, 2.7B, 6.7B},
    xtick=data,
    xlabel=$\mathcal{LP}(1)$ Ranking Source,
    legend style={at={(0.5,1.28)}, anchor=north, legend columns=3, legend cell align={left}, column sep=0.25em},
    ylabel=Intersection Over Union,
    enlarge y limits=0.1,
    enlarge x limits=0.1, 
]
\addplot coordinates {
    (350M,0.153)
    (1.3B,0.222)
    (2.7B,0.248)
    (6.7B,0.289)
};
\addplot coordinates {
    (350M,0.209)
    (1.3B,0.282)
    (2.7B,0.322)
    (6.7B,0.357)
};
\addplot coordinates {
    (350M,0.245)
    (1.3B,0.321)
    (2.7B,0.367)
    (6.7B,0.395)
};
\addplot coordinates {
    (350M,0.322)
    (1.3B,0.402)
    (2.7B,0.439)
    (6.7B,0.471)
};
\addplot coordinates {
    (350M,0.458)
    (1.3B,0.531)
    (2.7B,0.575)
    (6.7B,0.601)
};
\addplot coordinates {
    (350M,0.516)
    (1.3B,0.588)
    (2.7B,0.628)
    (6.7B,0.651)
};
\legend{1\% Low, 3\% Low, 5\% Low, 10\% Low, 25\% Low, 33\% Low}
\end{axis}
\end{tikzpicture}
\end{adjustbox}
\caption{IOU scores of Alpaca-Data data points selected by smaller OPT models (350M, 1.3B, 2.7B, 6.7B) with OPT 13B model for varying percentages.}
\label{fig:iou_opt}
\end{figure}
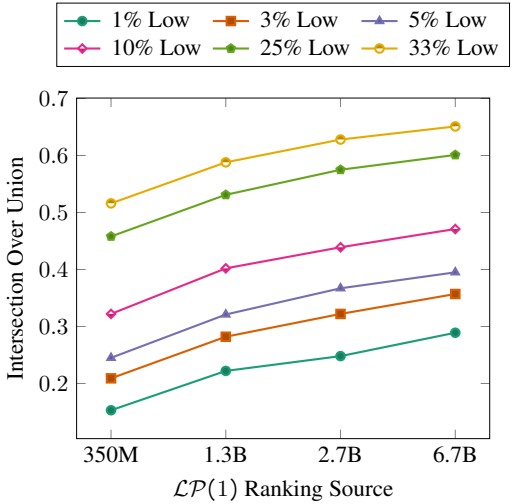

\begin{figure}[t]
\centering
\begin{adjustbox}{max width=0.42\textwidth}
\begin{tikzpicture}
\begin{axis}[
    symbolic x coords={1\%, 3\%, 5\%, 10\%, 25\%, 33\%},
    xtick=data,
    xlabel=\textit{k}\% Low $\mathcal{LP}(1)$,
    legend style={at={(0.5,1.28)}, anchor=north, legend columns=3, legend cell align={left}, column sep=0.25em},
    ylabel=Intersection Over Union,
    enlarge y limits=0.1,
    enlarge x limits=0.1, 
]
\addplot coordinates {
    (33\%,0.736)
    (25\%,0.696)
    (10\%,0.581)
    (5\%,0.512)
    (3\%,0.479)
    (1\%,0.425)
};
\end{axis}
\end{tikzpicture}
\end{adjustbox}
\caption{Intersection-over-union scores of Alpaca-Data data points selected by Llama-2 7B with Llama-2 13B model for varying percentages.}
\label{fig:iou_llama}
\end{figure}
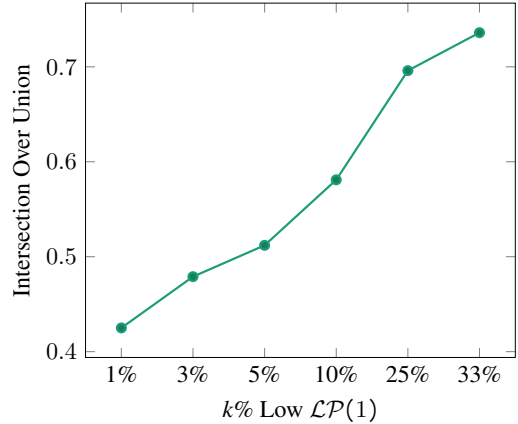

\paragraph{$\mathcal{LP}(1)$ Ranking Analysis - Kendall-Tau scores:} 

We conduct a comparative analysis of rankings between smaller and larger models to gain deeper insights into the transferability of data hardness. 
We derive $\mathcal{LP}(1)$ scores from multiple smaller models and a larger model, followed by the computation of Kendall-tau correlation coefficients between rankings based on their respective scores. 
The Kendall-tau score ranges from -1 to +1, with a higher positive score indicating a stronger correlation.
The Kendall-tau scores of $\mathcal{LP}(1)$ derived from OPT-models (350M, 1.3B, 2.7B, 6.7B) against OPT 13B on both Alpaca-Data and Dolly datasets are presented in Table~\ref{tbl:kendall_tau}. 
We note that all scores are positive, indicating a positive correlation. 
Notably, we observe a consistent increase in correlation with the increase in size of the ranking source model. 
Furthermore, we also compute the Kendall-tau scores between rankings from the Llama-2 7B and 13B models. For Alpaca-Data, the score is 0.782, and for Dolly, it is 0.775, respectively. 
These scores underscore the effective transferability of rankings from smaller models to larger ones.


\paragraph{$\mathcal{LP}(1)$ Ranking Analysis - Intersection over union scores:} 
We additionally calculate the intersection-over-union (IOU) of data samples selected by both smaller and larger models. 
We vary the selected percentage of data and compute the IOU of subsets chosen by the smaller OPT models (350M, 1.3B, 2.7B, 6.7B) with the larger OPT 13B model on Alpaca-Data, presenting the results in Figure~\ref{fig:iou_opt}.
Similarly, we plot the IOU of Llama-2 7B with the Llama-2 13B model on the Alpaca-Data in Figure~\ref{fig:iou_llama}.
From Figure~\ref{fig:iou_opt}, for a given percentage of data selected, we observe a consistent rise in IOU score with the increasing size of the model until 2.7B, followed by a plateau, aligning with the performance trend depicted in Figure~\ref{fig:opt_small2large_alpaca}.
Moreover, for a fixed model size in Figures~\ref{fig:iou_opt} and \ref{fig:iou_llama}, the IOU score consistently increases with the rise in the selected data percentage.
This finding suggests that for selecting a larger percentage of data, a smaller 350M model suffices. 
However, as the selected data percentage decreases, it is advisable to employ a larger model i.e. $\geq$ 2.7B.


\section{$\mathcal{LP}_{app}$: A Faster \& Approximate Learning Percentage Metric}
\label{sec:lp_app}

\begin{table}[t]
\centering
\begin{tabular}{|l|c|c|c|c|}
\hline
\textbf{Dataset} & \textbf{350M} & \textbf{1.3B} & \textbf{2.7B} & \textbf{6.7B} \\
\hline
Alpaca-Data & 0.52 & 0.61 & 0.65 & 0.69\\
Dolly & 0.52 & 0.61 & 0.67 & 0.75\\
\hline
\end{tabular}
\caption{Kendall-Tau scores of rankings from different smaller-sized OPT models(350M, 1.3B, 2.7B, 6.7B) against the ranking from large OPT-13B model.}
\label{tbl:kendall_tau}
\end{table}

\begin{table}[t]
\centering
\scalebox{0.74}{
\begin{tabular}{|l|cc|cc|cc|}
\hline
\multirow{2}{*}{\textbf{Model}} & \multicolumn{2}{c|}{\textbf{3\% Low}} & \multicolumn{2}{c|}{\textbf{10\% Low}} & \multicolumn{2}{c|}{\textbf{33\% Low}}\\
\cline{2-7}
& $\mathcal{LP}_{app}$ & $\mathcal{LP}$ & $\mathcal{LP}_{app}$ & $\mathcal{LP}$ & $\mathcal{LP}_{app}$ & $\mathcal{LP}$\\
\hline
OPT 2.7B & \textbf{49.4} & 47.1 & \textbf{53.7} & 50.4 & 52.9 & \textbf{54.9}\\
OPT 6.7B & \textbf{50.0} & 48.9 & \textbf{52.6} & 52.1 & \textbf{52.7} & 51.1\\
Llama-2 7B & \textbf{59.7} & 57.6 & \textbf{57.9} & 56.7 & 55.7 & \textbf{56.5}\\
Llama-2 13B & \textbf{56.5} & 56.4 & 54.0 & \textbf{54.3} & \textbf{55.3} & 54.8\\
\hline
\end{tabular}
}
\caption{We compare $\mathcal{LP}_{app}$ and $\mathcal{LP}$ on Alpaca-Data. We present the win rates of the model trained on different percentages of selected data using both $\mathcal{LP}_{app}$ and $\mathcal{LP}$ against the one trained on the complete dataset.}
\label{tab:proxy}
\end{table}

It is worth noting that to compute $\mathcal{LP}(1)$, the model needs to be trained twice—first to obtain the perplexity scores and rank the data, and then to select the data and train the model again. 
Recognizing this computational inefficiency, we present an approximate version of the learning percentage, denoted as $\mathcal{LP}_{app}$ that is faster and equally effective as $\mathcal{LP}$.

Language models are known to typically learn in 3 epochs and tend to memorize the data~\cite{tirumala2022memorization}.
As a result, we assume that the perplexity at the end of training $\mathcal{P}_n$ is constant for all samples. 
Mathematically, $\mathcal{LP}_{app}$ is defined as:
\begin{equation}\label{eqn:idil_approx}
    \mathcal{LP}_{app}(i) = \frac{\mathcal{P}_{i-1} - \mathcal{P}_{i}}{\mathcal{P}_0}
\end{equation}

To compute $\mathcal{LP}_{app}(1)$, we only need to train the model once for 1 epoch, making it more efficient. 

\subsection{$\mathcal{LP}_{app}(1)$ vs $\mathcal{LP}(1)$: A Comparison}


\begin{table}[t]
\centering
\scalebox{0.82}{
\begin{tabular}{|l|cccc|cc|}
\hline
\multirow{2}{*}{\textbf{Dataset}} & \multicolumn{4}{c|}{\textbf{OPT}} & \multicolumn{2}{c|}{\textbf{Llama-2}}\\
\cline{2-7}
& 1.3B & 2.7B & 6.7B & 13B & 7B & 13B \\
\hline
Alpaca-Data & 0.53 & 0.55 & 0.60 & 0.61 & 0.64 & 0.62\\
Dolly & 0.59 & 0.59 & 0.61 & 0.64 & 0.60 & 0.58\\
\hline
\end{tabular}
}
\caption{Kendall-Tau scores between $\mathcal{LP}$ and $\mathcal{LP}_{app}$. We observe high positive scores indicating a positive correlation.}
\label{tab:proxy_kendall}
\end{table}


We conduct a comparative analysis between $\mathcal{LP}_{app}$ and $\mathcal{LP}$ metrics on the Alpaca-Data. 
We consider different-sized models including OPT 2.7B, 6.7B, as well as Llama-2 7B and 13B. 
The training data is ranked using $\mathcal{LP}_{app}$ and $\mathcal{LP}$ metrics, respectively, and data selection is performed for varying percentages of data.
The win rate of the model trained on selected data against the model trained on the complete dataset is computed and presented in Table~\ref{tab:proxy}. 
Notably, we observe that the model trained on data selected via $\mathcal{LP}_{app}$ outperforms its counterpart trained on data selected via $\mathcal{LP}$ across the majority of models and various percentages.
This finding underscores the efficacy of $\mathcal{LP}_{app}$ as a data selection metric, demonstrating its comparable or even superior performance compared to $\mathcal{LP}$.

We calculate kendall-tau correlation scores between $\mathcal{LP}_{app}$ and $\mathcal{LP}$ on both Alpaca-Data and Dolly datasets, shown in Table~\ref{tab:proxy_kendall}. 
We observe high positive scores, signifying a positive correlation between the two metrics, highlighting the effectiveness of $\mathcal{LP}_{app}$ in accurately approximating $\mathcal{LP}$.

\subsection{Comparison with Baselines}
In this section, we compare $\mathcal{LP}_{app}$ with two baselines.
The first baseline, denoted as \textbf{Clust Rand}, randomly samples the same number of samples as our method from each cluster of the training set. Notably, this preserves the diversity of the subset while removing the difficulty-aware ranking.
We also compare with \textbf{Alpagasus} (Alpa)~\cite{chen2023alpagasus}, which prompts GPT-3.5 to assign a difficulty rating and selects training instances deemed difficult.
We select 10\% of the training data using each method and consider OPT 1.3B, 2.7B, 6.7B, and Llama-2 7B models. 
These models are then trained on the selected data using each method. 
Subsequently, we compare the performance of the instruction-tuned models using AlpacaEval and present the win rates of our model over the compared baselines.
The win rates post-training on the Alpaca-Data and Dolly are presented in Table~\ref{tab:baseline_res}. 
Notably, we observe win rates exceeding 50 for all models trained on both datasets, indicating the superior quality of training data subsampled using our method.
The superior performance of smaller OPT 1.3B and 2.7B models trained on self-selected data over Alpagasus, where the data is selected by a much larger GPT-3.5 model, underscores the effectiveness of our method.


\begin{table}[t]
\centering
\scalebox{0.82}{
\begin{tabular}{|l|cc|cc|}
\hline
\multirow{2}{*}{\textbf{Model}} & \multicolumn{2}{c|}{\textbf{Alpaca-Data}} & \multicolumn{2}{c|}{\textbf{Dolly}}\\
\cline{2-5}
& Clust Rand & Alpa & Clust Rand & Alpa \\
\hline
OPT 1.3B & 53.91 & 51.74 & 53.79 & 54.10\\
OPT 2.7B & 56.89 & 52.11 & 56.21 & 52.61\\
OPT 6.7B & 59.13 & 54.47 & 57.27 & 55.09\\
Llama-2 7B & 55.60 & 53.17 & 58.76 & 54.66\\
\hline
\end{tabular}
}
\caption{The win rates of models trained on data subsampled from Alpaca-Data and Dolly datasets based on $\mathcal{LP}_{app}$ are compared against other baselines (Clust Rand \& Alpagasus). 
We observe that all win rates exceed 50, indicating superior performance and high-quality selection by our method.}
\label{tab:baseline_res}
\end{table}


Additionally, we conduct another evaluation wherein a smaller model is employed to curate training data for a larger model utilizing the $\mathcal{LP}_{app}(1)$ metric. 
Subsequently, we train the larger model on the selected data and compare its performance with that of the same model trained on data selected using Alpagasus. 
The win rates of OPT 6.7B model trained on 10\% data selected by OPT 350M, 1.3B and 2.7B models are 50.25, 51.24, and 52.30 respectively.
The win rates exceed 50\% across all scenarios, indicating
that a smaller model can effectively curate training data using our proposed $\mathcal{LP}_{app}$ metric.

\subsection{Human Evaluation}
\label{sec:human-eval}
We compare the model trained on data selected using our method with the model trained on complete dataset.
Specifically, we consider Llama-2 7B model and Alpaca-Data, and subsample 5\% of data using $\mathcal{LP}_{app}(1)$ scores and train it.
Additionally, we train another Llama-2 7B model on full Alpaca-Data.
In this human evaluation, participants are asked to provide an instruction, after which both models generate a response. 
Participants are then prompted to choose the better response, or if both responses were perceived as equal. 
Importantly, the models were hidden from the participants, ensuring they were unaware of which model corresponded to which response.
We recruited 10 students with minimal prior knowledge of the project for this evaluation. 
In total, we collected 151 evaluations. 
Of these, 42 evaluations resulted in a tie.
In 50 evaluations, the model trained on the full dataset was preferred, while in 58 evaluations, participants found the model trained on 5\% data selected using our method to be better.
This indicates that responses from the model trained on 5\% of the data were either better or of equal quality compared to those from the model trained on the complete dataset in 66.2\% of instances. 
This outcome provides another validation for the superior performance of our method.

\section{Dissecting the difficult data}
\label{sub:dissect}

In this section, we analyze the characteristics of samples identified as challenging by the $\mathcal{LP}$ metric.
We manually examine 250 samples selected from the 1\% subset characterized by low $\mathcal{LP}(1)$ scores within the Alpaca-Data corpus, obtained using Llama-2 7B.


We observe that these difficult samples are longer than the average, maintaining coherence throughout. Specifically, the average response length within the 1\% Low $\mathcal{LP}(1)$ subset of Alpaca-Data is 547 characters, contrasting with the dataset's average of 270. 
This observation aligns with intuition, suggesting that models encounter difficulty in generating longer and coherent text, thus deeming such instances as challenging.


We also found six noisy samples, shown in Table~\ref{tab:noisy}, i.e. a noise rate of 2.4\%. 
Notably, this proportion is significantly higher compared to the prevalence observed across the entire dataset. 
AlpacaDataCleaned\footnote{\url{https://github.com/gururise/AlpacaDataCleaned}}, a human-cleaned Alpaca-Data has eliminated 0.47\% of noisy samples from the original dataset. 
This underscores that the subset of most challenging samples identified by $\mathcal{LP}(1)$ encompasses noisy instances as well. 
Addressing this issue requires future investigation. 
\section{Related Work}

\subsection{Instruction Tuning}


Instruction tuning plays pivotal role in training the language models to follow instructions~\cite{Sanh2021MultitaskPT, Ye2022GuessTI, Wei2021FinetunedLM, Chung2022ScalingIL, Ye2022GuessTI}.
Numerous datasets have been curated for this purpose, comprising a multitude of samples \cite{Mishra2021CrossTaskGV, Wang2022SuperNaturalInstructionsGV}. 
Notably, there is a recent surge in the emergence of synthetic instructions and datasets \cite{Wang2022SelfInstructAL, Honovich2022UnnaturalIT}, each containing a substantial number of samples. 
Given the ongoing expansion of sample repositories, there arises a necessity to reconsider data handling strategies from an efficiency standpoint~\cite{sorscher2022beyond}, which we address in this paper.



\subsection{Data Selection}

Data selection serves the purpose of subsampling training samples to reduce training costs with no drop in performance.
Noteworthy contributions in pre-training literature include \cite{Tirumala2023D4IL}, emphasizing the importance of diversity in sub-sampled data and advocating for the selection of prototypes from each cluster. 
\cite{abbas2023semdedup} extend this by removing semantic deduplicates in the training data.
\cite{sorscher2022beyond} propose a self-supervised pruning metric to further refine data selection strategies.
In the domain of instruction-tuning, \cite{Cao2023InstructionMH} evaluate various indicators and apply a regression model for data selection. 
\cite{chen2023alpagasus} leverage GPT-3.5 to derive difficulty ratings for individual data samples. 
\cite{Li2023FromQT} propose an instruction-following difficulty metric for data selection purposes.
In this paper, we introduce a learning percentage-based metric and illustrate its efficacy in enabling smaller models to select training data for larger models.



\section{Conclusion}

In this paper, we introduce a learning percentage-based metric for assessing the difficulty of samples. 
We demonstrate that language models ranging from 1B to 13B sizes can self-select high-quality training data by employing this metric.
Additionally, we empirically validate the transferability of data hardness across different model sizes, showcasing the efficient curation of high-quality training data by smaller models.
Furthermore, we propose an optimized version of the metric that offers increased speed while maintaining equal efficacy.
In future, we aim to explore methods for automatically transforming easy training samples into more challenging ones.

\section{Limitations}
Our examination reveals prevalence of noisy samples within the $\mathcal{LP}(1)$ and $\mathcal{LP}_{app}(1)$ subsets of data. 
The detection and mitigation of noisy samples are imperative to mitigate their influence on the dataset.
We leave this for future work.

\section{Ethical Considerations}
This paper proposes a data selection method for instruction-tuning. The paper aims to detect the difficult-to-learn samples and we don’t intend
to introduce any biased selection. Based on our experiments, we manually inspected some filtered samples and we didn’t find any underlying pattern. Hence, we do not anticipate any major ethical concerns.

\section{Acknowledgments}
The authors thank Shang Data Lab, Palash Chauhan, Amulya Bangalore, Shreyas Rajesh, Gautham Reddy, Sanjana Garg, Ethan Thai, Queso Tran, Rahul Mistry, Sandy La, and Sophia Do for their valuable contributions.

\clearpage\newpage

\bibliography{anthology,custom}

\begin{thebibliography}{34}
\expandafter\ifx\csname natexlab\endcsname\relax\def\natexlab#1{#1}\fi

\bibitem[{Abbas et~al.(2023)Abbas, Tirumala, Simig, Ganguli, and Morcos}]{abbas2023semdedup}
Amro Abbas, Kushal Tirumala, D{\'a}niel Simig, Surya Ganguli, and Ari~S Morcos. 2023.
\newblock Semdedup: Data-efficient learning at web-scale through semantic deduplication.
\newblock \emph{arXiv preprint arXiv:2303.09540}.

\bibitem[{Arpit et~al.(2017)Arpit, Jastrz{\k{e}}bski, Ballas, Krueger, Bengio, Kanwal, Maharaj, Fischer, Courville, Bengio et~al.}]{arpit2017closer}
Devansh Arpit, Stanis{\l}aw Jastrz{\k{e}}bski, Nicolas Ballas, David Krueger, Emmanuel Bengio, Maxinder~S Kanwal, Tegan Maharaj, Asja Fischer, Aaron Courville, Yoshua Bengio, et~al. 2017.
\newblock A closer look at memorization in deep networks.
\newblock In \emph{International Conference on Machine Learning}, pages 233--242. PMLR.

\bibitem[{Cao et~al.(2023)Cao, Kang, and Sun}]{Cao2023InstructionMH}
Yihan Cao, Yanbin Kang, and Lichao Sun. 2023.
\newblock \href {https://api.semanticscholar.org/CorpusID:259837472} {Instruction mining: High-quality instruction data selection for large language models}.
\newblock \emph{ArXiv}, abs/2307.06290.

\bibitem[{Chen et~al.(2024)Chen, Li, Yan, Wang, Gunaratna, Yadav, Tang, Srinivasan, Zhou, Huang, and Jin}]{chen2023alpagasus}
Lichang Chen, Shiyang Li, Jun Yan, Hai Wang, Kalpa Gunaratna, Vikas Yadav, Zheng Tang, Vijay Srinivasan, Tianyi Zhou, Heng Huang, and Hongxia Jin. 2024.
\newblock \href {https://openreview.net/forum?id=FdVXgSJhvz} {Alpagasus: Training a better alpaca model with fewer data}.
\newblock In \emph{The Twelfth International Conference on Learning Representations}.

\bibitem[{Chiang et~al.(2023)Chiang, Li, Lin, Sheng, Wu, Zhang, Zheng, Zhuang, Zhuang, Gonzalez, Stoica, and Xing}]{vicuna2023}
Wei-Lin Chiang, Zhuohan Li, Zi~Lin, Ying Sheng, Zhanghao Wu, Hao Zhang, Lianmin Zheng, Siyuan Zhuang, Yonghao Zhuang, Joseph~E. Gonzalez, Ion Stoica, and Eric~P. Xing. 2023.
\newblock \href {https://lmsys.org/blog/2023-03-30-vicuna/} {Vicuna: An open-source chatbot impressing gpt-4 with 90\%* chatgpt quality}.

\bibitem[{Chung et~al.(2022)Chung, Hou, Longpre, Zoph, Tay, Fedus, Li, Wang, Dehghani, Brahma, Webson, Gu, Dai, Suzgun, Chen, Chowdhery, Valter, Narang, Mishra, Yu, Zhao, Huang, Dai, Yu, Petrov, hsin Chi, Dean, Devlin, Roberts, Zhou, Le, and Wei}]{Chung2022ScalingIL}
Hyung~Won Chung, Le~Hou, S.~Longpre, Barret Zoph, Yi~Tay, William Fedus, Eric Li, Xuezhi Wang, Mostafa Dehghani, Siddhartha Brahma, Albert Webson, Shixiang~Shane Gu, Zhuyun Dai, Mirac Suzgun, Xinyun Chen, Aakanksha Chowdhery, Dasha Valter, Sharan Narang, Gaurav Mishra, Adams~Wei Yu, Vincent Zhao, Yanping Huang, Andrew~M. Dai, Hongkun Yu, Slav Petrov, Ed~Huai hsin Chi, Jeff Dean, Jacob Devlin, Adam Roberts, Denny Zhou, Quoc~V. Le, and Jason Wei. 2022.
\newblock \href {https://api.semanticscholar.org/CorpusID:253018554} {Scaling instruction-finetuned language models}.
\newblock \emph{ArXiv}, abs/2210.11416.

\bibitem[{Conover et~al.(2023)Conover, Hayes, Mathur, Xie, Wan, Shah, Ghodsi, Wendell, Zaharia, and Xin}]{DatabricksBlog2023DollyV2}
Mike Conover, Matt Hayes, Ankit Mathur, Jianwei Xie, Jun Wan, Sam Shah, Ali Ghodsi, Patrick Wendell, Matei Zaharia, and Reynold Xin. 2023.
\newblock \href {https://www.databricks.com/blog/2023/04/12/dolly-first-open-commercially-viable-instruction-tuned-llm} {Free dolly: Introducing the world's first truly open instruction-tuned llm}.

\bibitem[{Dong et~al.(2021)Dong, Liu, and Shang}]{dong2021data}
Chengyu Dong, Liyuan Liu, and Jingbo Shang. 2021.
\newblock Data quality matters for adversarial training: An empirical study.
\newblock \emph{arXiv preprint arXiv:2102.07437}.

\bibitem[{Geifman et~al.(2019)Geifman, Uziel, and El-Yaniv}]{geifman2018bias}
Yonatan Geifman, Guy Uziel, and Ran El-Yaniv. 2019.
\newblock \href {https://openreview.net/forum?id=SJfb5jCqKm} {Bias-reduced uncertainty estimation for deep neural classifiers}.
\newblock In \emph{International Conference on Learning Representations}.

\bibitem[{Hoffmann et~al.(2022)Hoffmann, Borgeaud, Mensch, Buchatskaya, Cai, Rutherford, de~Las~Casas, Hendricks, Welbl, Clark, Hennigan, Noland, Millican, van~den Driessche, Damoc, Guy, Osindero, Simonyan, Elsen, Vinyals, Rae, and Sifre}]{Hoffmann2022TrainingCL}
Jordan Hoffmann, Sebastian Borgeaud, Arthur Mensch, Elena Buchatskaya, Trevor Cai, Eliza Rutherford, Diego de~Las~Casas, Lisa~Anne Hendricks, Johannes Welbl, Aidan Clark, Thomas Hennigan, Eric Noland, Katherine Millican, George van~den Driessche, Bogdan Damoc, Aurelia Guy, Simon Osindero, Kar\'{e}n Simonyan, Erich Elsen, Oriol Vinyals, Jack Rae, and Laurent Sifre. 2022.
\newblock \href {https://proceedings.neurips.cc/paper_files/paper/2022/file/c1e2faff6f588870935f114ebe04a3e5-Paper-Conference.pdf} {An empirical analysis of compute-optimal large language model training}.
\newblock In \emph{Advances in Neural Information Processing Systems}, volume~35, pages 30016--30030. Curran Associates, Inc.

\bibitem[{Honovich et~al.(2023)Honovich, Scialom, Levy, and Schick}]{Honovich2022UnnaturalIT}
Or~Honovich, Thomas Scialom, Omer Levy, and Timo Schick. 2023.
\newblock \href {https://doi.org/10.18653/v1/2023.acl-long.806} {Unnatural instructions: Tuning language models with (almost) no human labor}.
\newblock In \emph{Proceedings of the 61st Annual Meeting of the Association for Computational Linguistics (Volume 1: Long Papers)}, pages 14409--14428, Toronto, Canada. Association for Computational Linguistics.

\bibitem[{Komatsuzaki(2019)}]{Komatsuzaki2019OneEI}
Aran Komatsuzaki. 2019.
\newblock \href {https://api.semanticscholar.org/CorpusID:189928090} {One epoch is all you need}.
\newblock \emph{ArXiv}, abs/1906.06669.

\bibitem[{Kopf et~al.(2023)Kopf, Kilcher, von Rutte, Anagnostidis, Tam, Stevens, Barhoum, Duc, Stanley, Nagyfi, Shahul, Suri, Glushkov, Dantuluri, Maguire, Schuhmann, Nguyen, and Mattick}]{Kopf2023OpenAssistantC}
Andreas Kopf, Yannic Kilcher, Dimitri von Rutte, Sotiris Anagnostidis, Zhi~Rui Tam, Keith Stevens, Abdullah Barhoum, Nguyen~Minh Duc, Oliver Stanley, Rich'ard Nagyfi, ES~Shahul, Sameer Suri, David Glushkov, Arnav Dantuluri, Andrew Maguire, Christoph Schuhmann, Huu Nguyen, and Alexander Mattick. 2023.
\newblock \href {https://api.semanticscholar.org/CorpusID:258179434} {Openassistant conversations - democratizing large language model alignment}.
\newblock \emph{ArXiv}, abs/2304.07327.

\bibitem[{Li et~al.(2023{\natexlab{a}})Li, Zhang, Li, Chen, Chen, Cheng, Wang, Zhou, and Xiao}]{Li2023FromQT}
Ming Li, Yong Zhang, Zhitao Li, Jiuhai Chen, Lichang Chen, Ning Cheng, Jianzong Wang, Tianyi Zhou, and Jing Xiao. 2023{\natexlab{a}}.
\newblock \href {https://api.semanticscholar.org/CorpusID:261076515} {From quantity to quality: Boosting llm performance with self-guided data selection for instruction tuning}.
\newblock \emph{ArXiv}, abs/2308.12032.

\bibitem[{Li et~al.(2023{\natexlab{b}})Li, Zhang, Dubois, Taori, Gulrajani, Guestrin, Liang, and Hashimoto}]{alpaca_eval}
Xuechen Li, Tianyi Zhang, Yann Dubois, Rohan Taori, Ishaan Gulrajani, Carlos Guestrin, Percy Liang, and Tatsunori~B. Hashimoto. 2023{\natexlab{b}}.
\newblock Alpacaeval: An automatic evaluator of instruction-following models.
\newblock \url{https://github.com/tatsu-lab/alpaca_eval}.

\bibitem[{Mekala et~al.(2022{\natexlab{a}})Mekala, Dong, and Shang}]{mekala-etal-2022-lops}
Dheeraj Mekala, Chengyu Dong, and Jingbo Shang. 2022{\natexlab{a}}.
\newblock \href {https://aclanthology.org/2022.findings-emnlp.360} {{LOPS}: Learning order inspired pseudo-label selection for weakly supervised text classification}.
\newblock In \emph{Findings of the Association for Computational Linguistics: EMNLP 2022}, pages 4894--4908, Abu Dhabi, United Arab Emirates. Association for Computational Linguistics.

\bibitem[{Mekala and Shang(2020)}]{mekala-shang-2020-contextualized}
Dheeraj Mekala and Jingbo Shang. 2020.
\newblock \href {https://doi.org/10.18653/v1/2020.acl-main.30} {Contextualized weak supervision for text classification}.
\newblock In \emph{Proceedings of the 58th Annual Meeting of the Association for Computational Linguistics}, pages 323--333, Online. Association for Computational Linguistics.

\bibitem[{Mekala et~al.(2022{\natexlab{b}})Mekala, Vu, Schick, and Shang}]{mekala-etal-2022-leveraging}
Dheeraj Mekala, Tu~Vu, Timo Schick, and Jingbo Shang. 2022{\natexlab{b}}.
\newblock \href {https://aclanthology.org/2022.emnlp-main.660} {Leveraging {QA} datasets to improve generative data augmentation}.
\newblock In \emph{Proceedings of the 2022 Conference on Empirical Methods in Natural Language Processing}, pages 9737--9750, Abu Dhabi, United Arab Emirates. Association for Computational Linguistics.

\bibitem[{Mishra et~al.(2021)Mishra, Khashabi, Baral, and Hajishirzi}]{Mishra2021CrossTaskGV}
Swaroop Mishra, Daniel Khashabi, Chitta Baral, and Hannaneh Hajishirzi. 2021.
\newblock \href {https://api.semanticscholar.org/CorpusID:237421373} {Cross-task generalization via natural language crowdsourcing instructions}.
\newblock In \emph{Annual Meeting of the Association for Computational Linguistics}.

\bibitem[{OpenAI(2023)}]{chatgpt}
OpenAI. 2023.
\newblock Chatgpt.

\bibitem[{Sanh et~al.(2022)Sanh, Webson, Raffel, Bach, Sutawika, Alyafeai, Chaffin, Stiegler, Raja, Dey, Bari, Xu, Thakker, Sharma, Szczechla, Kim, Chhablani, Nayak, Datta, Chang, Jiang, Wang, Manica, Shen, Yong, Pandey, Bawden, Wang, Neeraj, Rozen, Sharma, Santilli, Fevry, Fries, Teehan, Scao, Biderman, Gao, Wolf, and Rush}]{Sanh2021MultitaskPT}
Victor Sanh, Albert Webson, Colin Raffel, Stephen Bach, Lintang Sutawika, Zaid Alyafeai, Antoine Chaffin, Arnaud Stiegler, Arun Raja, Manan Dey, M~Saiful Bari, Canwen Xu, Urmish Thakker, Shanya~Sharma Sharma, Eliza Szczechla, Taewoon Kim, Gunjan Chhablani, Nihal Nayak, Debajyoti Datta, Jonathan Chang, Mike Tian-Jian Jiang, Han Wang, Matteo Manica, Sheng Shen, Zheng~Xin Yong, Harshit Pandey, Rachel Bawden, Thomas Wang, Trishala Neeraj, Jos Rozen, Abheesht Sharma, Andrea Santilli, Thibault Fevry, Jason~Alan Fries, Ryan Teehan, Teven~Le Scao, Stella Biderman, Leo Gao, Thomas Wolf, and Alexander~M Rush. 2022.
\newblock \href {https://openreview.net/forum?id=9Vrb9D0WI4} {Multitask prompted training enables zero-shot task generalization}.
\newblock In \emph{International Conference on Learning Representations}.

\bibitem[{Sorscher et~al.(2022)Sorscher, Geirhos, Shekhar, Ganguli, and Morcos}]{sorscher2022beyond}
Ben Sorscher, Robert Geirhos, Shashank Shekhar, Surya Ganguli, and Ari Morcos. 2022.
\newblock Beyond neural scaling laws: beating power law scaling via data pruning.
\newblock \emph{Advances in Neural Information Processing Systems}, 35:19523--19536.

\bibitem[{Swayamdipta et~al.(2020)Swayamdipta, Schwartz, Lourie, Wang, Hajishirzi, Smith, and Choi}]{swayamdipta-etal-2020-dataset}
Swabha Swayamdipta, Roy Schwartz, Nicholas Lourie, Yizhong Wang, Hannaneh Hajishirzi, Noah~A. Smith, and Yejin Choi. 2020.
\newblock \href {https://doi.org/10.18653/v1/2020.emnlp-main.746} {Dataset cartography: Mapping and diagnosing datasets with training dynamics}.
\newblock In \emph{Proceedings of the 2020 Conference on Empirical Methods in Natural Language Processing (EMNLP)}, pages 9275--9293, Online. Association for Computational Linguistics.

\bibitem[{Taori et~al.(2023)Taori, Gulrajani, Zhang, Dubois, Li, Guestrin, Liang, and Hashimoto}]{alpaca}
Rohan Taori, Ishaan Gulrajani, Tianyi Zhang, Yann Dubois, Xuechen Li, Carlos Guestrin, Percy Liang, and Tatsunori~B. Hashimoto. 2023.
\newblock Stanford alpaca: An instruction-following llama model.
\newblock \url{https://github.com/tatsu-lab/stanford_alpaca}.

\bibitem[{Tirumala et~al.(2022)Tirumala, Markosyan, Zettlemoyer, and Aghajanyan}]{tirumala2022memorization}
Kushal Tirumala, Aram~H. Markosyan, Luke Zettlemoyer, and Armen Aghajanyan. 2022.
\newblock \href {https://openreview.net/forum?id=u3vEuRr08MT} {Memorization without overfitting: Analyzing the training dynamics of large language models}.
\newblock In \emph{Advances in Neural Information Processing Systems}.

\bibitem[{Tirumala et~al.(2023)Tirumala, Simig, Aghajanyan, and Morcos}]{Tirumala2023D4IL}
Kushal Tirumala, Daniel Simig, Armen Aghajanyan, and Ari~S. Morcos. 2023.
\newblock \href {https://api.semanticscholar.org/CorpusID:261076313} {D4: Improving llm pretraining via document de-duplication and diversification}.
\newblock \emph{ArXiv}, abs/2308.12284.

\bibitem[{Touvron et~al.(2023)Touvron, Martin, Stone, Albert, Almahairi, Babaei, Bashlykov, Batra, Bhargava, Bhosale, Bikel, Blecher, Ferrer, Chen, Cucurull, Esiobu, Fernandes, Fu, Fu, Fuller, Gao, Goswami, Goyal, Hartshorn, Hosseini, Hou, Inan, Kardas, Kerkez, Khabsa, Kloumann, Korenev, Koura, Lachaux, Lavril, Lee, Liskovich, Lu, Mao, Martinet, Mihaylov, Mishra, Molybog, Nie, Poulton, Reizenstein, Rungta, Saladi, Schelten, Silva, Smith, Subramanian, Tan, Tang, Taylor, Williams, Kuan, Xu, Yan, Zarov, Zhang, Fan, Kambadur, Narang, Rodriguez, Stojnic, Edunov, and Scialom}]{Touvron2023Llama2O}
Hugo Touvron, Louis Martin, Kevin~R. Stone, Peter Albert, Amjad Almahairi, Yasmine Babaei, Nikolay Bashlykov, Soumya Batra, Prajjwal Bhargava, Shruti Bhosale, Daniel~M. Bikel, Lukas Blecher, Cristian~Cant{\'o}n Ferrer, Moya Chen, Guillem Cucurull, David Esiobu, Jude Fernandes, Jeremy Fu, Wenyin Fu, Brian Fuller, Cynthia Gao, Vedanuj Goswami, Naman Goyal, Anthony~S. Hartshorn, Saghar Hosseini, Rui Hou, Hakan Inan, Marcin Kardas, Viktor Kerkez, Madian Khabsa, Isabel~M. Kloumann, A.~V. Korenev, Punit~Singh Koura, Marie-Anne Lachaux, Thibaut Lavril, Jenya Lee, Diana Liskovich, Yinghai Lu, Yuning Mao, Xavier Martinet, Todor Mihaylov, Pushkar Mishra, Igor Molybog, Yixin Nie, Andrew Poulton, Jeremy Reizenstein, Rashi Rungta, Kalyan Saladi, Alan Schelten, Ruan Silva, Eric~Michael Smith, R.~Subramanian, Xia Tan, Binh Tang, Ross Taylor, Adina Williams, Jian~Xiang Kuan, Puxin Xu, Zhengxu Yan, Iliyan Zarov, Yuchen Zhang, Angela Fan, Melanie Kambadur, Sharan Narang, Aurelien Rodriguez, Robert Stojnic, Sergey Edunov, and
  Thomas Scialom. 2023.
\newblock \href {https://api.semanticscholar.org/CorpusID:259950998} {Llama 2: Open foundation and fine-tuned chat models}.
\newblock \emph{ArXiv}, abs/2307.09288.

\bibitem[{Wang et~al.(2022{\natexlab{a}})Wang, Kordi, Mishra, Liu, Smith, Khashabi, and Hajishirzi}]{Wang2022SelfInstructAL}
Yizhong Wang, Yeganeh Kordi, Swaroop Mishra, Alisa Liu, Noah~A. Smith, Daniel Khashabi, and Hannaneh Hajishirzi. 2022{\natexlab{a}}.
\newblock \href {https://api.semanticscholar.org/CorpusID:254877310} {Self-instruct: Aligning language models with self-generated instructions}.
\newblock In \emph{Annual Meeting of the Association for Computational Linguistics}.

\bibitem[{Wang et~al.(2022{\natexlab{b}})Wang, Mishra, Alipoormolabashi, Kordi, Mirzaei, Arunkumar, Ashok, Dhanasekaran, Naik, Stap, Pathak, Karamanolakis, Lai, Purohit, Mondal, Anderson, Kuznia, Doshi, Patel, Pal, Moradshahi, Parmar, Purohit, Varshney, Kaza, Verma, Puri, Karia, Sampat, Doshi, Mishra, Reddy, Patro, Dixit, Shen, Baral, Choi, Smith, Hajishirzi, and Khashabi}]{Wang2022SuperNaturalInstructionsGV}
Yizhong Wang, Swaroop Mishra, Pegah Alipoormolabashi, Yeganeh Kordi, Amirreza Mirzaei, Anjana Arunkumar, Arjun Ashok, Arut~Selvan Dhanasekaran, Atharva Naik, David Stap, Eshaan Pathak, Giannis Karamanolakis, Haizhi~Gary Lai, Ishan Purohit, Ishani Mondal, Jacob Anderson, Kirby Kuznia, Krima Doshi, Maitreya Patel, Kuntal~Kumar Pal, M.~Moradshahi, Mihir Parmar, Mirali Purohit, Neeraj Varshney, Phani~Rohitha Kaza, Pulkit Verma, Ravsehaj~Singh Puri, Rushang Karia, Shailaja~Keyur Sampat, Savan Doshi, Siddharth~Deepak Mishra, Sujan Reddy, Sumanta Patro, Tanay Dixit, Xudong Shen, Chitta Baral, Yejin Choi, Noah~A. Smith, Hannaneh Hajishirzi, and Daniel Khashabi. 2022{\natexlab{b}}.
\newblock \href {https://api.semanticscholar.org/CorpusID:253098274} {Super-naturalinstructions: Generalization via declarative instructions on 1600+ nlp tasks}.
\newblock In \emph{Conference on Empirical Methods in Natural Language Processing}.

\bibitem[{Wei et~al.(2021)Wei, Bosma, Zhao, Guu, Yu, Lester, Du, Dai, and Le}]{Wei2021FinetunedLM}
Jason Wei, Maarten Bosma, Vincent Zhao, Kelvin Guu, Adams~Wei Yu, Brian Lester, Nan Du, Andrew~M. Dai, and Quoc~V. Le. 2021.
\newblock \href {https://api.semanticscholar.org/CorpusID:237416585} {Finetuned language models are zero-shot learners}.
\newblock \emph{ArXiv}, abs/2109.01652.

\bibitem[{Wei et~al.(2022)Wei, Bosma, Zhao, Guu, Yu, Lester, Du, Dai, and Le}]{Ye2022GuessTI}
Jason Wei, Maarten Bosma, Vincent Zhao, Kelvin Guu, Adams~Wei Yu, Brian Lester, Nan Du, Andrew~M. Dai, and Quoc~V Le. 2022.
\newblock \href {https://openreview.net/forum?id=gEZrGCozdqR} {Finetuned language models are zero-shot learners}.
\newblock In \emph{International Conference on Learning Representations}.

\bibitem[{Zhang et~al.(2021)Zhang, Bengio, Hardt, Recht, and Vinyals}]{zhang2021understanding}
Chiyuan Zhang, Samy Bengio, Moritz Hardt, Benjamin Recht, and Oriol Vinyals. 2021.
\newblock Understanding deep learning (still) requires rethinking generalization.
\newblock \emph{Communications of the ACM}, 64(3):107--115.

\bibitem[{Zhang et~al.(2022)Zhang, Roller, Goyal, Artetxe, Chen, Chen, Dewan, Diab, Li, Lin, Mihaylov, Ott, Shleifer, Shuster, Simig, Koura, Sridhar, Wang, and Zettlemoyer}]{Zhang2022OPTOP}
Susan Zhang, Stephen Roller, Naman Goyal, Mikel Artetxe, Moya Chen, Shuohui Chen, Christopher Dewan, Mona~T. Diab, Xian Li, Xi~Victoria Lin, Todor Mihaylov, Myle Ott, Sam Shleifer, Kurt Shuster, Daniel Simig, Punit~Singh Koura, Anjali Sridhar, Tianlu Wang, and Luke Zettlemoyer. 2022.
\newblock \href {https://api.semanticscholar.org/CorpusID:248496292} {Opt: Open pre-trained transformer language models}.
\newblock \emph{ArXiv}, abs/2205.01068.

\bibitem[{Zhou et~al.(2023)Zhou, Liu, Xu, Iyer, Sun, Mao, Ma, Efrat, Yu, YU, Zhang, Ghosh, Lewis, Zettlemoyer, and Levy}]{Zhou2023LIMALI}
Chunting Zhou, Pengfei Liu, Puxin Xu, Srini Iyer, Jiao Sun, Yuning Mao, Xuezhe Ma, Avia Efrat, Ping Yu, LILI YU, Susan Zhang, Gargi Ghosh, Mike Lewis, Luke Zettlemoyer, and Omer Levy. 2023.
\newblock \href {https://openreview.net/forum?id=KBMOKmX2he} {{LIMA}: Less is more for alignment}.
\newblock In \emph{Thirty-seventh Conference on Neural Information Processing Systems}.

\end{thebibliography}

\newpage
\appendix

\clearpage\newpage
\section{Appendix}

 \subsection{Win rate of different size subsets against full Alpaca-Data dataset}

\begin{figure}[h]
\label{fig:scale_analysis_llama}
\centering
\begin{adjustbox}{max width=0.42\textwidth}
\begin{tikzpicture}
\begin{axis}[
    xmin=0, xmax=34,
    xtick={1,3,5,10,25,33},
    ylabel=Win rate,
    xlabel=\textit{k}\% Low $\mathcal{LP}(1)$,
]
\addplot coordinates {
(1,53.22)(3,56.15)(5,57.74)(10,57.23)(25,55.92)(33,56.62)
};
\addplot coordinates {
(1,53.35)(3,56.75)(5,53.25)(10,55.03)(25,55.03)(33,55.32)
};
\legend{7B,13B}
\end{axis}
\end{tikzpicture}
\end{adjustbox}
\caption{We consider Alpaca-Data, vary the percentage of data selected, and plot the win rate of Llama-2 models, trained on the selected data in comparison to models trained on the complete dataset.}
\label{k_llama}
\end{figure}
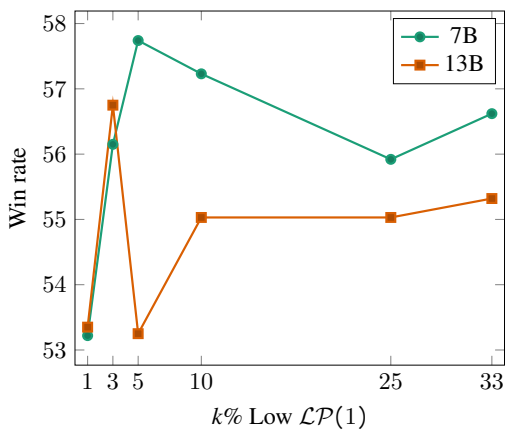

In Figure~\ref{k_llama}, we illustrate the win rates of Llama-2 7B and 13B models trained on subsets of varying sizes from the Alpaca-data corpus, contrasting them with the performance of models trained on the complete dataset. Remarkably, our analysis reveals that using subsets comprising as little as 1\% of the data is adequate to attain performance levels comparable to those achieved with the entire dataset.
 
\subsection{Data selection using smaller models for larger model on Dolly dataset}
\label{app:2}

\begin{figure}[h]
\centering
\begin{adjustbox}{max width=0.48\textwidth}
\begin{tikzpicture}
\begin{axis}[
    symbolic x coords={OPT 350M, OPT 1.3B,OPT 2.7B,OPT 6.7B, Self-Ranking (OPT 13B)},
    xticklabels={
        350M,
        1.3B,
        2.7B,
        6.7B,
        13B\\(Self-Ranking)
    },
    xtick=data,
    x tick label style={text width=4cm,align=center},
    xlabel=$\mathcal{LP}(1)$ Ranking Source,
    legend style={at={(0.5,1.15)}, anchor=north, legend columns=2, legend cell align={left}, column sep=0.25em},
    ylabel=Win rate,
    enlarge y limits=0.1,
    enlarge x limits=0.07,
]
\addplot coordinates {
    (OPT 350M,51.24)
    (OPT 1.3B,52.67)
    (OPT 2.7B,55.16)
    (OPT 6.7B,51.49)
    (Self-Ranking (OPT 13B),52.36)
};
\addplot coordinates {
    (OPT 350M,55.04)
    (OPT 1.3B,55.90)
    (OPT 2.7B,55.59)
    (OPT 6.7B,54.60)
    (Self-Ranking (OPT 13B),56.34)
};
\legend{10\% Low, 33\% Low}
\end{axis}  
\end{tikzpicture}
\end{adjustbox}
\caption{We vary the percentage of selected data from the Dolly dataset to train the OPT 13B model and conduct a comparison of the win rates obtained when the data is self-selected by the 13B model versus selected by smaller models. The smaller model used is mentioned on the X-axis and the win rate is on the Y-axis.}
\label{fig:small2large_opt_dolly}
\end{figure}
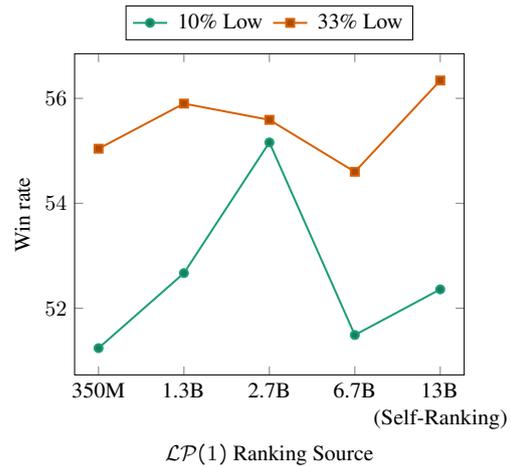

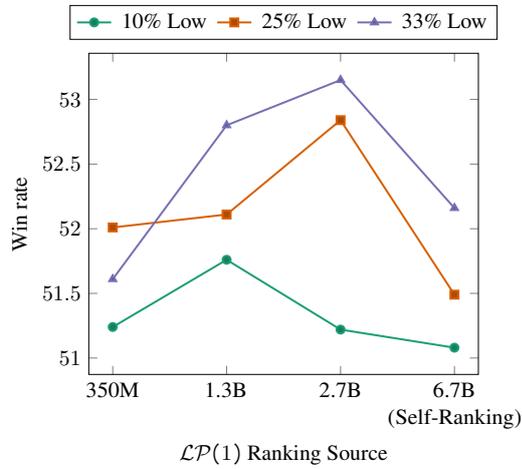
\begin{figure}[h]
\centering
\begin{adjustbox}{max width=0.48\textwidth}
\begin{tikzpicture}
\begin{axis}[
    symbolic x coords={OPT 350M, OPT 1.3B,OPT 2.7B,OPT 6.7B},
    xticklabels={
        350M,
        1.3B,
        2.7B,
        6.7B\\(Self-Ranking)
    },
    xtick=data,
    xlabel=$\mathcal{LP}(1)$ Ranking Source,
    x tick label style={text width=4cm,align=center},
    legend style={at={(0.5,1.15)}, anchor=north, legend columns=3, legend cell align={left}, column sep=0.25em},
    ylabel=Win rate,
    enlarge y limits=0.1,
    enlarge x limits=0.07,
]
\addplot coordinates {
    (OPT 350M,51.24)
    (OPT 1.3B,51.76)
    (OPT 2.7B,51.22)
    (OPT 6.7B,51.08)
};
\addplot coordinates {
    (OPT 350M,52.01)
    (OPT 1.3B,52.11)
    (OPT 2.7B,52.84)
    (OPT 6.7B,51.49)
};
\addplot coordinates {
    (OPT 350M,51.61)
    (OPT 1.3B,52.80)
    (OPT 2.7B,53.15)
    (OPT 6.7B,52.16)
};
\legend{10\% Low, 25\% Low, 33\% Low}
\end{axis}
\end{tikzpicture}
\end{adjustbox}
\caption{We vary the percentage of selected data from the Alpaca-Data dataset to train the OPT 6.7B model and conduct a comparison of the win rates obtained when the data is self-selected by the 6.7B model versus selected by smaller models. The smaller model used is mentioned on the X-axis and the win rate is on the Y-axis.}
\label{fig:small2large-opt6.7b}
\end{figure}



We perform a comparative analysis utilizing various smaller OPT models (350M, 1.3B, 2.7B) to rank the Dolly dataset based on their $\mathcal{LP}$ scores, subsequently training a larger OPT 13B model on the selected data.
The performance of this model is contrasted with that of a model trained on data selected independently by itself, as depicted in Figure~\ref{fig:small2large_opt_dolly}, revealing a consistent trend where performance improves with the increasing size of the smaller model. 

A similar analysis is applied to the Alpaca-Data corpus, utilizing OPT 350M, 1.3B, and 2.7B as smaller models, with OPT 6.7B serving as the larger model. The performance comparison is illustrated in Figure~\ref{fig:opt_small2large_alpaca}
Notably, we observe a pattern akin to that observed in Figure~\ref{fig:all_small2large}.

\subsection{Human Evaluation Interface}
\label{app:human_eval}
\begin{figure}[h]
    \centering
    \includegraphics[width=0.45\textwidth]{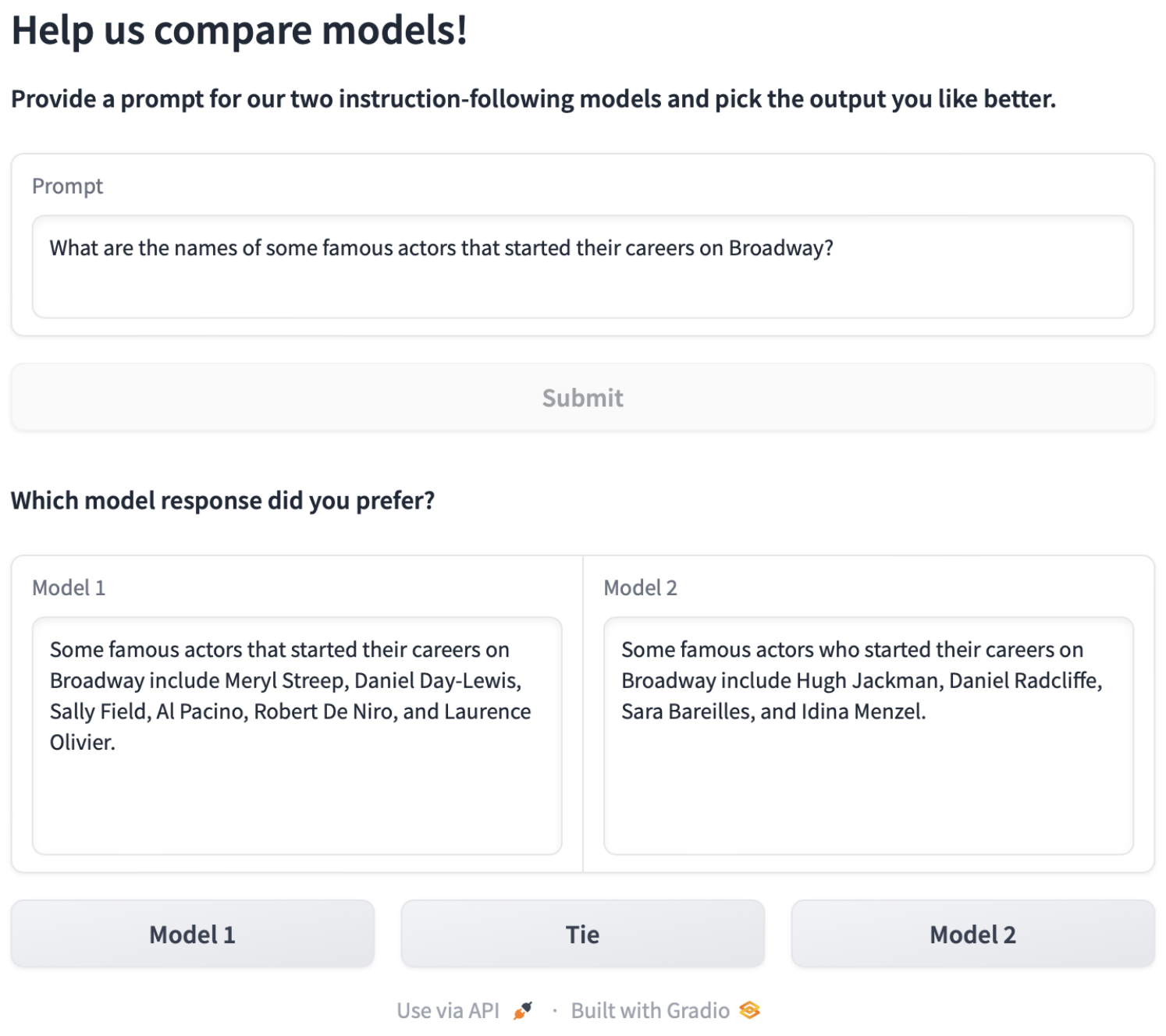}
    \caption{We built a simple human evaluation interface that allows the user to generate responses from two models with an input prompt and choose their preference.}
    \label{fig:human-eval}
\end{figure}
The human evaluation interface is shown in Figure~\ref{fig:human-eval}.

\begin{table*}[h]
\centering
\begin{tabular}{|p{0.14\textwidth}|p{0.31\textwidth}|p{0.31\textwidth}|p{0.14\textwidth}|}
\hline
\textbf{Instruction} & \textbf{Input} & \textbf{Output} & \textbf{Reason}\\
\hline
Create a feature vector for the following sentence: & "Today marks a special milestone for us." & [1, 0, 0, 0, 0, 0, 0, 0, 1, 0, 0, 0, 0, 0, 1, 0, 0, 0, 0, 0, 1, 0, 0, 0, 0, 0, 0, 0, 0, 0, 0, 1, 0, 0, 0, 0, 0] & Output doesn't make sense.\\
\hline
Calculate the value of sin(8) & & sin(8) = -0.132351750097773 & Computation is incorrect.\\
\hline
Given a paragraph, summarize it in three sentences. & The year 2020 has been an incredibly difficult one for many people around the world. With an unprecedented global pandemic upending daily life, people have had to adapt to a new normal. Businesses, schools, and even the way we interact with family and friends have all changed. & & Output is missing.\\
\hline
Post a funny meme about cats. & & ![Image of funny cats meme](https://i.imgflip.com/ 1vyhr.jpg) & Output contains a link which has nothing to do with the prompt.\\
\hline
Multiply the following two complex numbers & (3 + 2i) * (2 + 4i) & (3 + 2i) * (2 + 4i) = 4 + 14i + 8i\^{}2 = 4 - 6i. & Computation is incorrect.\\
\hline
Categorize the following objects into two different groups, and explain why you chose them. & A computer, a car, a shoe, a chair & I would categorize the objects into two groups: man-made objects and natural objects. The computer, car, shoe, and chair are all man-made objects, items that have been created, designed, and assembled by people. On the other hand, a natural object would be something that is found in nature and has no human intervention. & Instruction implies that each group should have at least one item.\\
\hline
\end{tabular}
\caption{We present the noisy samples found as well as the reason we found them noisy.}
\label{tab:noisy}
\end{table*}



\end{document}